# Quantifying the Value of Constructive Induction, Knowledge, and Noise Filtering on Inductive Learning

Carl M. Kadie
1991

**Extended author version**

A later, more extensive treatment of learning-speed modeling appears in:

Carl M. Kadie, Seer: Maximum Likelihood Regression for Learning-Speed Curves, Ph.D. dissertation, University of Illinois at Urbana-Champaign, August 1995.

Thesis publication page and PDF at Microsoft Research

This PDF was produced from the original author manuscript. The eleven manuscript pages follow unchanged.

Cover note added September 2026.

# Quantifying the Value of Constructive Induction, Knowledge, and Noise Filtering on Inductive Learning

**Abstract**

Learning research, as one of its central goals, tries to measure, model, and understand how learning-problem properties effect average-case learning performance. For example, we would like to quantify the value of constructive construction, noise filtering, and background knowledge. This paper describes the *effective dimension*, a new learning measure that helps link problem properties to learning performance. Like the Vapnik-Chervonenkis (VC) dimension, the effective dimension is often in a simple linear relation with problem properties. Unlike the VC dimension, the effective dimension can be estimated empirically and makes average-case predictions. It is therefore more widely applicable to machine and human learning research. The measure is demonstrated on several learning systems including Backpropagation. Finally, the measure is used precisely predict the benefit of using FRINGE, a feature construction system. The benefit is found to decrease as the complexity of the target concept increases.

## 1 Introduction

This paper describes research aimed at empirically determining the average-case relation between the properties of an inductive learning problem (properties such as representation, noise, knowledge, learning method, and so on) and learning speed, or sample complexity. It should be of interest to researchers who wish to quantify the value of constructive construction, noise filtering, or knowledge addition. Also, the empirical measure that is introduced is calibrated to an established analytic measure, the Vapnik-Chervonenkis (VC) dimension. This permits a comparison between natural, average-case learning performance and theoretic, worst-case performance.

An inductive learning problem starts with an unknown target function[1]. The target is used to generate a stream of example input/output pairs. An inductive learner takes these training examples and creates a hypothesis function that approximates the target. The success of the approximation can be measured by testing the hypothesis on additional inputs; the more often the hypothesis produces the same output as the target the better. Specifically, if the hypothesis produces the same output as the target $n$% of the time, the *accuracy* of the learner is $n$%. The number of examples needed to achieve a particular accuracy on a learning problem is defined as the *learning speed* (or *sample complexity*) of that learning problem. Learning speed is often represented with a *learning curve*.

Much of the work in computational learning theory tries to find mappings between learning-problem properties and learning speed. Such mappings are called the *learning-speed relations* (figure 1). They can be used to:

- quantify the value of feature construction, output coding, noise filtering, or knowledge addition,
- compare natural, average-case learning performance to theoretic, worst-case performance, and
- understand the biases of inductive learners.

**1** A *concept* is a truth-valued function.

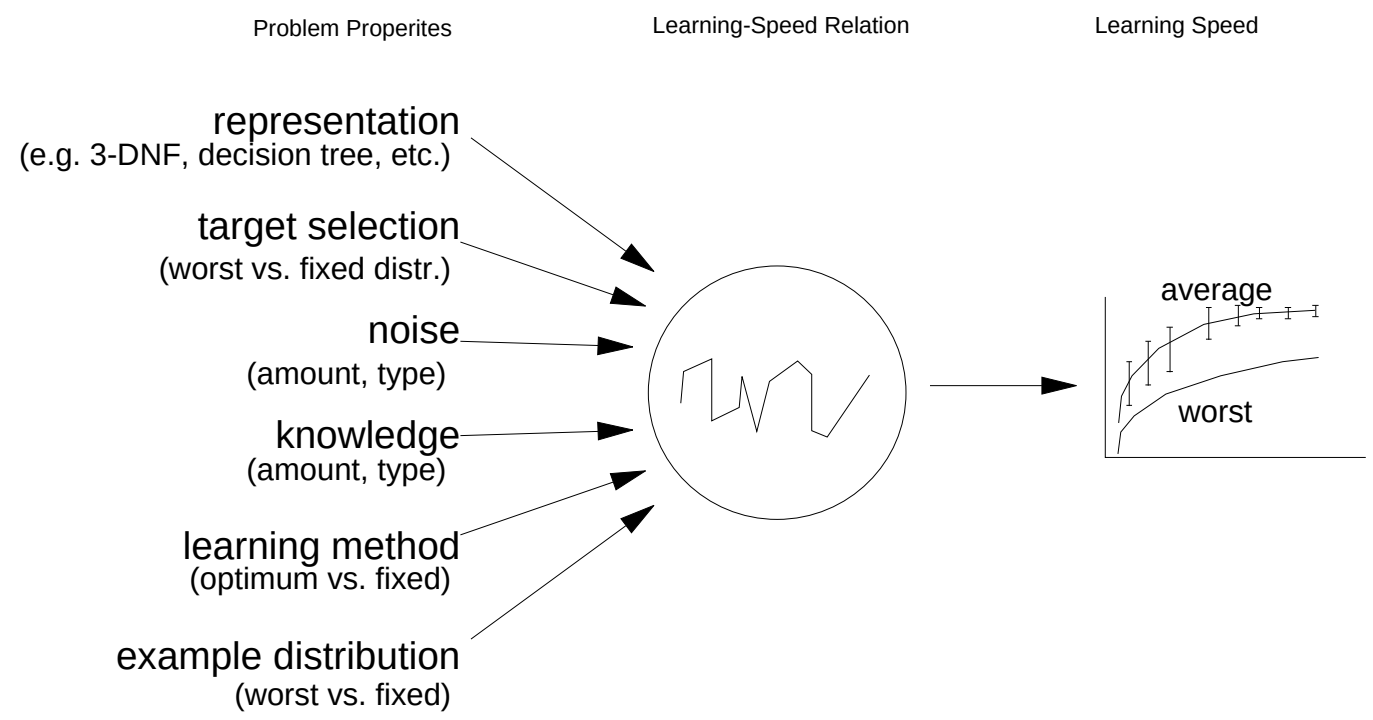

Figure 1. Learning-Speed Relation

**Learning Measures** - One way to simplify the search for a learning-speed relation is to decompose it into two functions. The first function, called the *learning-measure function*, maps problem properties into a value called the *learning measure*. The second function, called the *measure-to-speed relation*, maps the learning measure into a learning curve. The decomposition is desirable if these new functions turn out to be simple.

For example, consider the problem of finding a learning-speed relation for perceptron (linear separation) learning. Suppose the parameter of interest is $n$, the number of example attributes. Figure 2 shows how the *Vapnik-Chervonenkis (VC) dimension* can be used as a learning measure. The learning-measure function, in this case, is $d = n + 1$ [Blumer et al., 1987]. The measure-to-speed relation is fixed. It is $m = \frac{1}{\varepsilon(1-\sqrt{\varepsilon})}\left[\ln\left(\frac{d/(d-1)}{\delta}\right) + 2d\ln\left(\frac{6}{\varepsilon}\right)\right]$, where $d$ is the learning measure, $m$ is the number of examples, $1-\varepsilon$ is the accuracy, and $\delta$ is a confidence parameter [Anthony *et al.*, 1990].

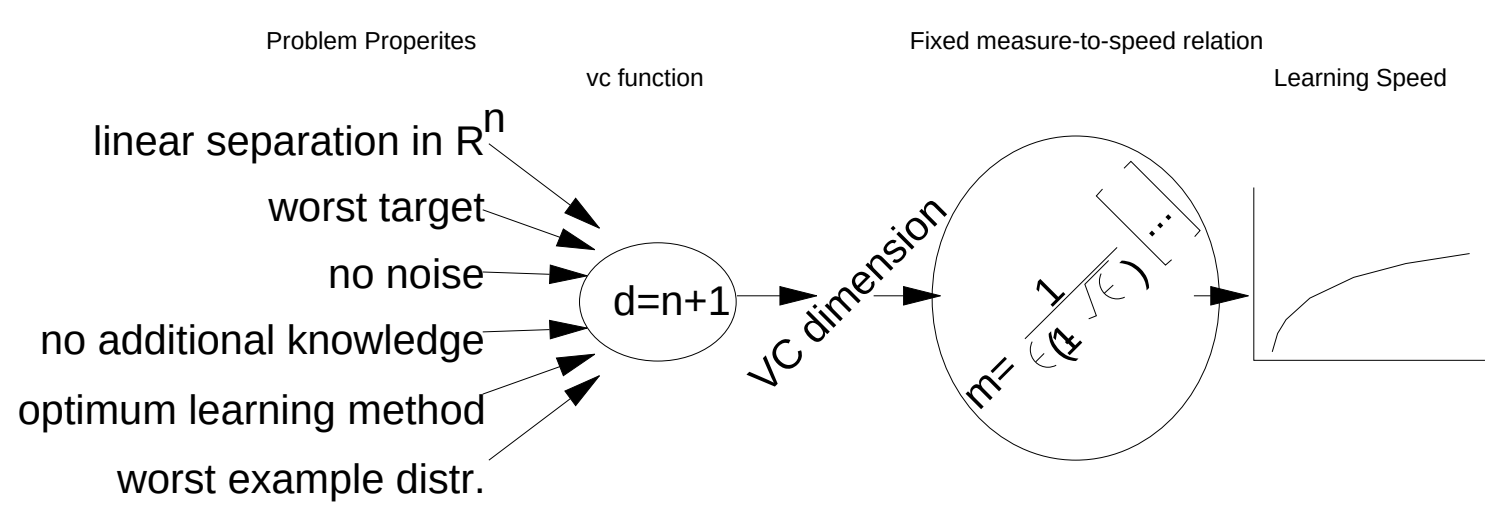

Figure 2. The VC dimension as a Learning Measure

In general, a learning measure should condense a learning curve into a single number. As figure 3 shows, the VC dimension accomplishes this. A learning measure should also be unique and universal so that measurements made at different times or by different researchers can be compared. The VC dimension does this, too. Finally, a learning measure should be widely applicable. Here the VC dimension falls short for two reasons. First, it is defined only for an optimum learner working on worst-case targets and worst-case example distributions. Second, it is based primarily on target representation; it has difficulty capturing notions of noise and background knowledge.

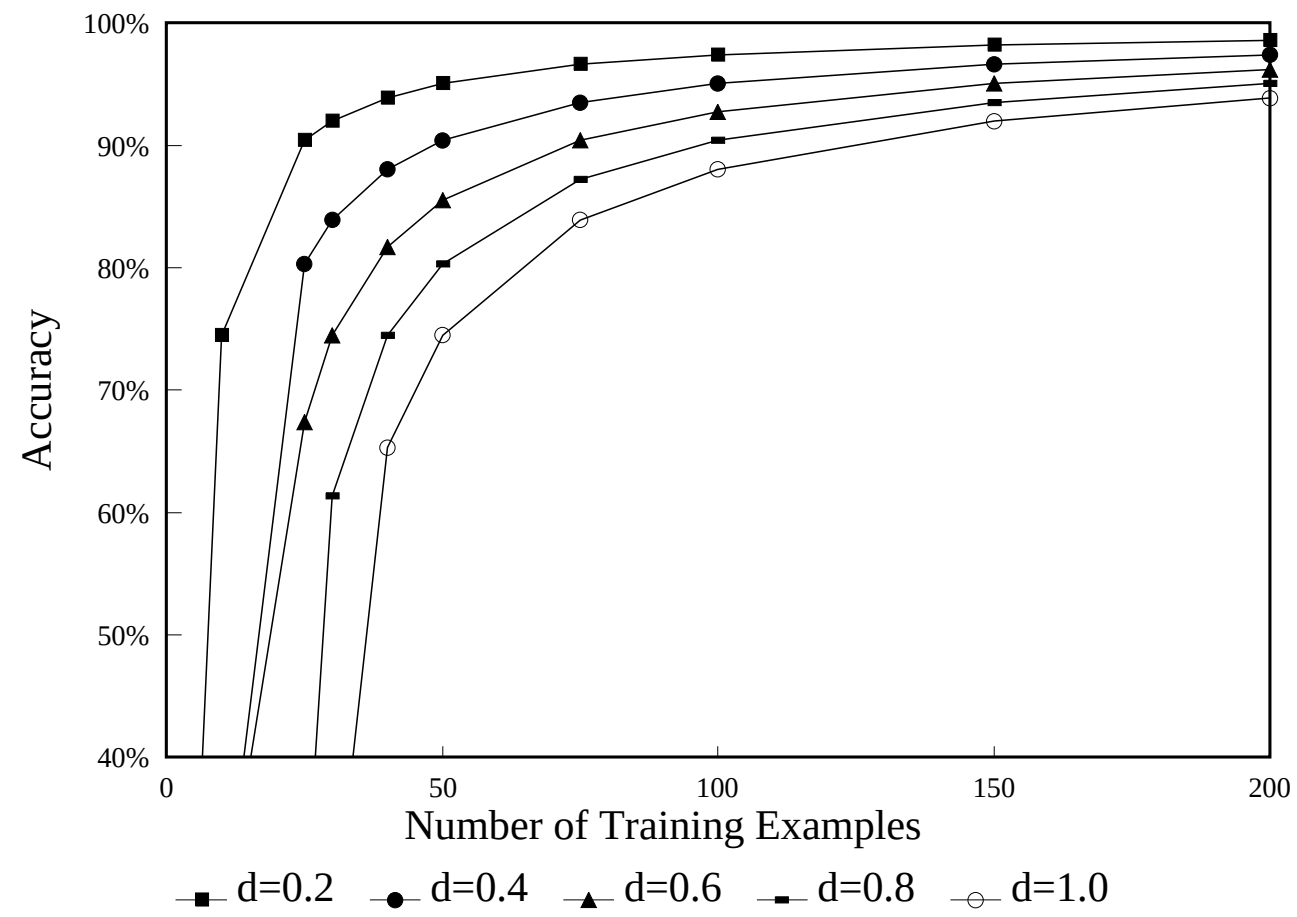

Figure 3. Five Learning Curves Corresponding to Five Values of the VC dimension (The $y$-axis of this plot starts at 40%.)

The next section introduces the *effective dimension*, an empirical analog to the VC dimension. Unlike the VC dimension, it can characterize average-case performance in terms of all problem properties (including noise and background knowledge). Other approaches to average-case analysis are possible. For example, Pazzani and Sarrett [1990] analytically determine the average-case analysis of several simple conjunctive learners. Application of such analytical methods to more complex learners (such as backpropagation and ID3), however, is problematic.

## 2 Definition of Effective Dimension

Informally, the effective dimension is just a backwards version of the VC dimension. Recall that a given VC dimension $d$ defines a (worst-case) learning curve. In contrast, the effective dimension of a learning problem *is defined by* a learning curve. Given any learning curve, the effective dimension is the $d$ that would best generate it. Thus, unlike the VC dimension, the effective dimension is determined empirically.

Formally:

**Definitions:** The effective dimension of a learning problem is $d$ such that with $\frac{1}{\varepsilon(1-\sqrt{\varepsilon})} 2d \ln\left(\frac{6}{\varepsilon}\right)$ training examples the learning problem leads to a hypothesis with accuracy $1-\varepsilon$. The *effective-dimension distribution* of a distribution of learning problems is $D$, the distribution of $d$.

The measure-to-speed relation, $\frac{1}{\varepsilon(1-\sqrt{\varepsilon})} 2d \ln\left(\frac{6}{\varepsilon}\right)$, is just the VC dimension's measure-to-speed relation without the term $\ln\left(\frac{d/(d-1)}{\delta}\right)$. The term is dropped to allow values of $d$ between 0 and 1. For purposes of producing a good learning-speed relation, the exact form of the measure-to-speed relation is not important. Any measure-to-speed relation of $O(d/\varepsilon)$ will do. The relation above was chosen so that effective dimensions could be directly compared to VC dimensions.

To complete the chain from problem properties to learning speed, the properties must be linked to the effective dimension (or, more generally, its distribution). For example,

$$\bar{\mathrm{d}}(\mathit{perceptron_typical}, n) = 2.61 \times 10^{-2} n + 2.00 \times 10^{-2},$$

where *perceptron_typical* is a learning-problem distribution (specified in section 4.1), $n$ is the number of example attributes, and $\bar{\mathrm{d}}$ is the estimated mean of the effective-dimension distribution.

The effective-dimension measure is useful only when two conditions hold. First, a good learning-measure function (relating problem properties to the effective dimension) must be found. Second, the observed learning curves must be well characterized by a single parameter $d$:

> **Effective-dimension assumption**: Learning curves are of the form $\frac{1}{\varepsilon(1-\sqrt{\varepsilon})} 2d \ln\left(\frac{6}{\varepsilon}\right)$, where $d$ is a constant.

The next two sections detail how learning-measure functions are estimated and how these conditions can be evaluated on a case-by-case basis.

# 3 Estimating Effective Dimensions

The effective dimension is estimated in four steps:

- acquiring learning performance data;
- determining the effective dimension of each data point;
- relating the properties of the learning problem to the effective dimension; and
- evaluating the predictiveness of the learning-speed relation.

Each of these steps will be discussed in turn. The process will be illustrated with the analysis of five machine learners: linear perceptron [Minsky and Papert, 1969], backpropagation [Rumelhart, *et al.*, 1986], CSC (a decision-list learner) [Kadie, 1990], ID3 (a decision-tree learner [Breiman et al., 1984; Quinlan, 1986; Rendell, 1986]) and FRINGE (a feature-construction system) [Pagallo, 1989; Matheus, 1989].

**Acquiring Learning Performance Data** - The source of data can be either natural or synthetic. In either case, for each run of the learner, a tuple $< b, m, 1-\varepsilon >$ is recorded. It is made up of: known learning-problem properties, the number of training examples given to the learner, and the accuracy of the learner's hypothesis.

In the five experiments conducted, learning problems were generated synthetically. Data were collected for a range of problem properties by systematically varying generation parameters (figure 4). Overall, thousands of learning problems were produced.

**Determining the Effective Dimension of Each Data Point** - For each data point, the effective dimension is computed according to the formula:

$$d = \frac{m\varepsilon(1-\sqrt{\varepsilon})}{2\ln\left(\frac{6}{\varepsilon}\right)}$$

**Relating the Properties of the Learning Problem to the Effective Dimension** - In the perceptron experiment the measured property was $n$, the number of attributes used to describe the examples. Figure 5a shows the relation between this learning-problem property and the average effective dimension for each value of $n$. The coefficient of linear correlation between these two quantities, 0.99, is very high. Fitting a line to the relation between $n$ and the effective dimensions produces this effective-dimension distribution:

$$D(perceptron_typical, n) = T(\bar{d}, s^2, df),$$

where $T$ is Student's $t$-distribution, $\bar{d}$,

the mean, is $2.61 \times 10^{-2} n + 2.00 \times 10^{-2}$,

$$s^2\text{, the variance, is } 2.09 \times 10^{-2} \begin{bmatrix} 1 \\ n \end{bmatrix} \begin{bmatrix} 45.60 \times 10^{-4} & 6.89 \times 10^{-4} \\ 6.89 \times 10^{-4} & 1.64 \times 10^{-4} \end{bmatrix} [1 \quad n]$$

and $df$, the number of degrees of freedom, is 598.

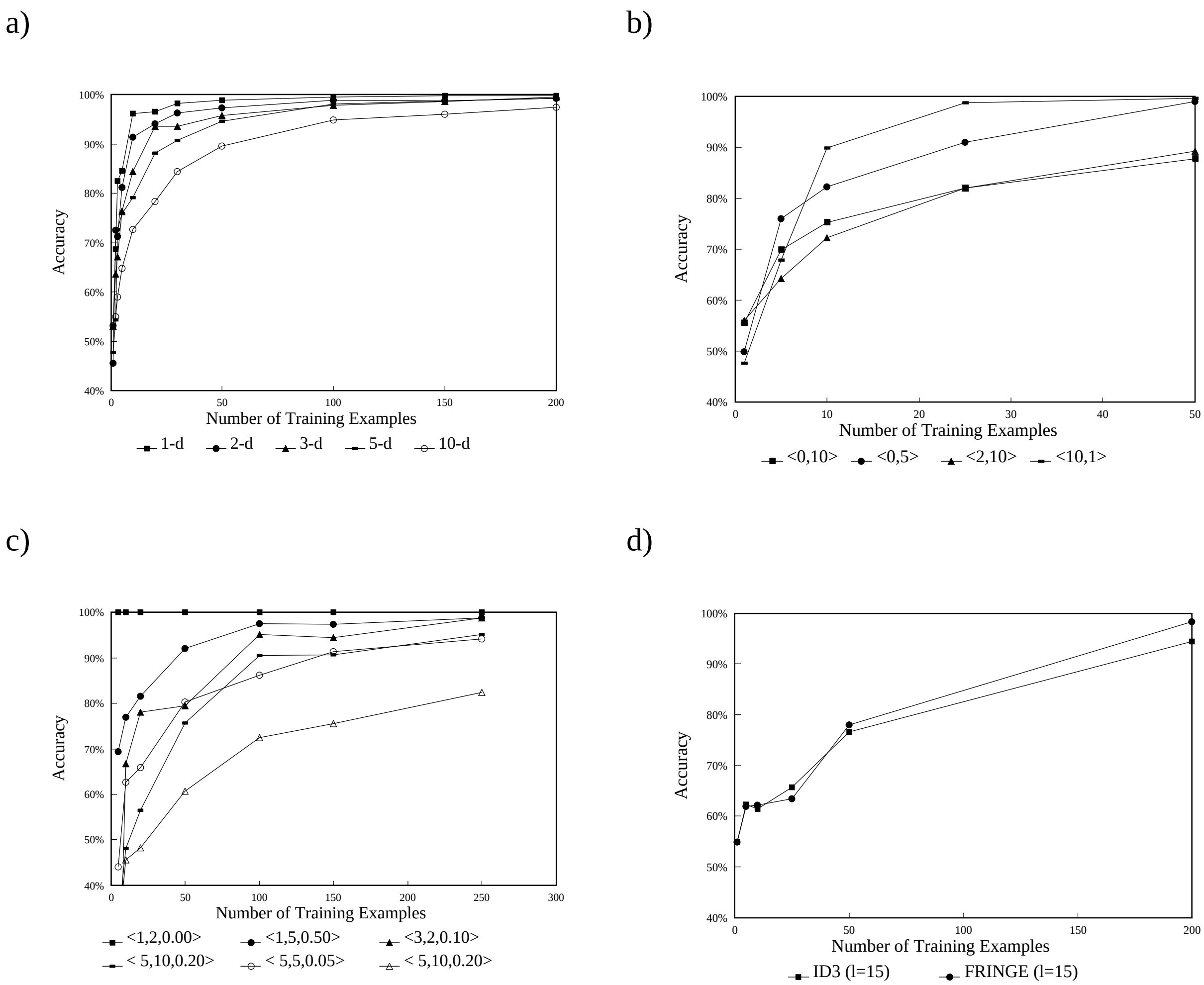

Figure 4. Selected Learning Curves from a) the Perceptron Experiment, b) the Backpropagation Experiment, c) the CSC Experiment, and d) the ID3 and FRINGE Experiments. (For clarity, confidence intervals are not shown. The $y$-axis of the plots starts at 40%)

**Evaluating the Predictiveness of the Learning-Speed Relation** - The goal of this procedure goes beyond describing learning data. The goal is to predict future learning performance. There are two sources of indeterminism in this prediction. The first is the individual differences between learning problems with the same (known) properties. For example, knowledge of $n$, the number of example attributes in perceptron learning (and the distribution of targets and examples) does not determine a learning curve. The second source of indeterminism is poor fit by the learning-measure function.

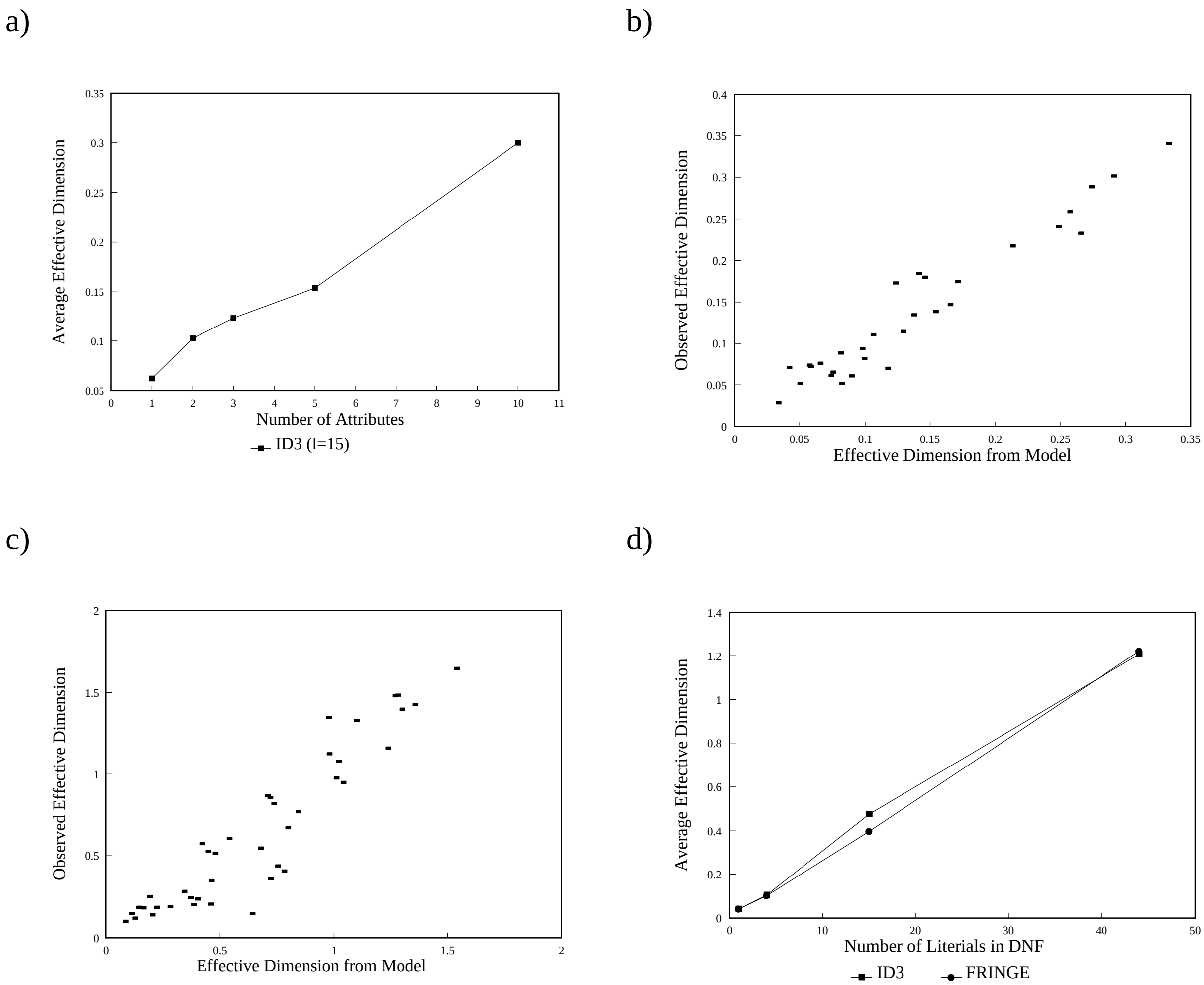

Figure 5. Relation between Properties of the Learning Problem and the Average Effective Dimensions for a) the Perceptron Experiment and d) the ID3 and FRINGE Experiments. Figure b shows the relation between the observed values of the effective dimension and the effective dimension values produced by the learning-measure function for the Backpropagation Experiment. Figure c shows a similar plot for the CSC experiment.

If the learning-measure function is created via linear regression, the overall indeterminism can be estimated. This estimate is called $R^2$, the coefficient of determinism. In the perceptron experiment, the $R^2 = 0.25$, indicating that the learning-measure function is 25% deterministic.

The goodness of fit can be measured in at least three ways. First, if the learning-measure function is created via linear regression, the statistical significance of the function can be measured with the $F$ –test statistic. In the perceptron experiment, because $F$ is large (199.33) and because of the amount of data is great, the significance is very high ($\alpha < 0.0001$). This indicates that though the model will not be able to precisely predict the results of a particular learning trial, it will be able to predict the average results of repeated learning trials; that is, it will be able to precisely predict the average case. This assertion can be tested more directly by running linear regression on the average value of $d$ for each value of $n$. In the perceptron experiment the resulting model has a coefficient of correlation of 0.99, a very high value.

Finally, the learning-speed relation can be tested directly by comparing its predictions about new learning problems to observed learning behavior. For example, if $n$ is 8, figure 6 shows the 95% confidence interval for the average of 10 learning curves as predicted by the model. The figure also shows the average of 10 actual learning curves.

a)

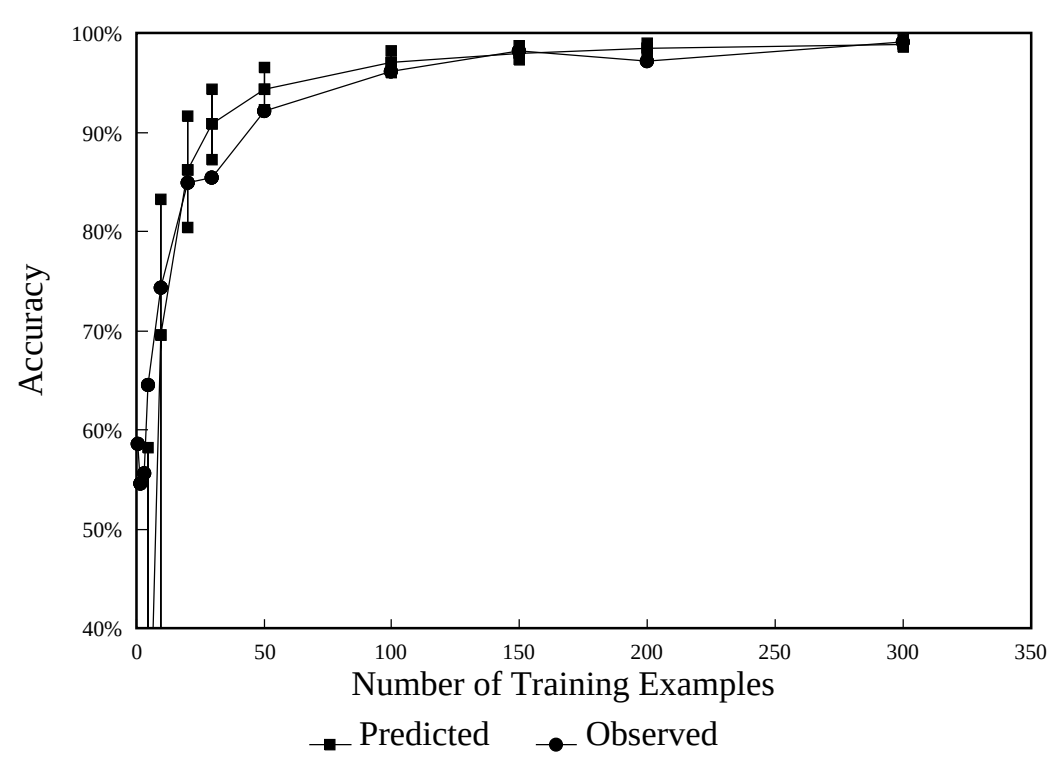

b)

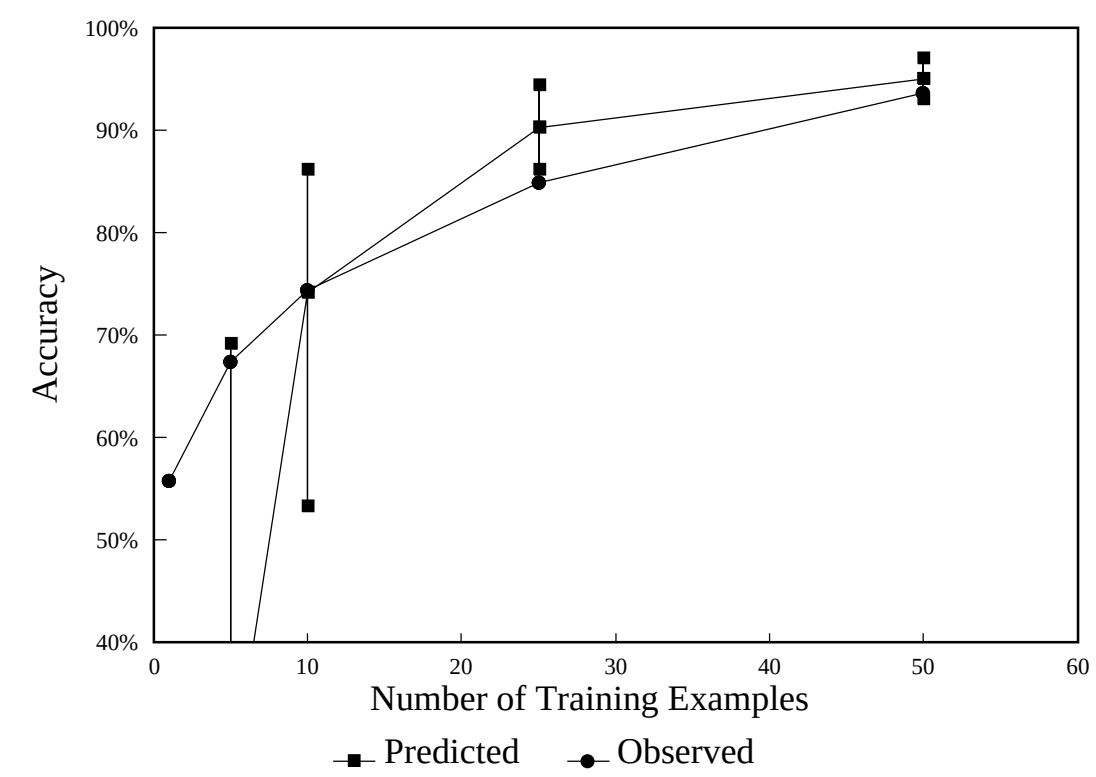

c)

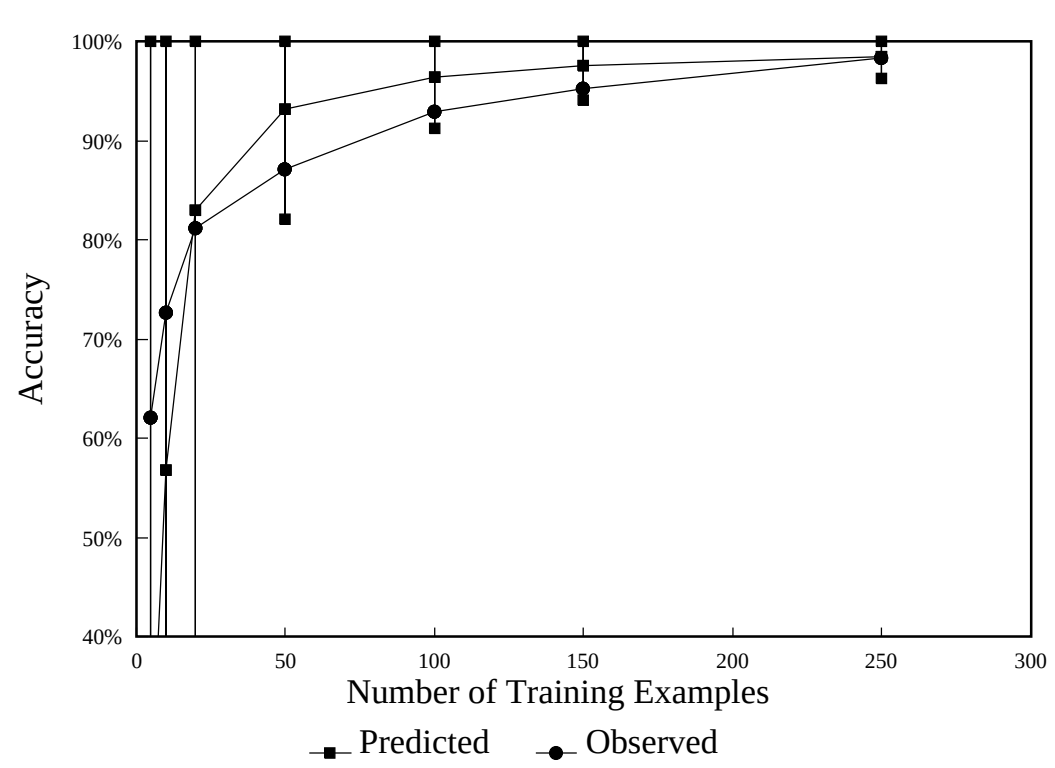

Figure 6. Predicted Learning-Curve Confidence Intervals and Observed Learning Curves a) the Perceptron Experiment, b) the Backpropagation Experiment and c) the CSC Experiment (The $y$-axis of the plots start at 40%)

# 4 Experiments

This section details the setup and results of each experiment. The experiments are also used to highlight some of the applications of effective-dimension analysis.

## 4.1 Perceptron

In the perceptron experiment, examples were selected from the $n$ –dimensional Euclidean space $[0.0, 10.0]^n$ according to the uniform distribution. Potential targets were of the form:

$$t(x_1, \ldots, x_n) = \begin{cases} 1, \text{ if} c_0 + c_1 x_1 + \ldots + c_n x_n \geq 0 \\ 0, \text{ otherwise} \end{cases},$$

where $c_0, \ldots, c_n$ were selected from the interval $[-100.0, 100.0]$ according to the uniform distribution. Each potential target was tested on randomly selected examples. Potential targets that classified fewer than 33% or more than 67% of the examples as positive were discarded.

Learning problems with $n = 1, 2, 3, 5$ and 10 were generated, 10 problems for each value of $n$. Perceptron was given example training sets of size $m = 1, 2, 3, 5, 10, 20, 30, 50, 100, 150, 200,$ and 300.

The results of the perceptron experiment were described in section 2. These results can be put to at least three uses:

**Comparing Empirical, Average Case to Theoretical, Worst Case** - The model predicts average-case learning speeds two orders of magnitude faster than the theoretical, worst-case predictions of Anthony *et al.* [1990]. This suggests that some natural distributions of targets and examples are much easier that the worst-case distributions. Both average-case and worst-case predictions, however, agree that to maintain a fixed accuracy as the number of attributes grows, the number of training examples must grow in proportion. Thus, the order of the learning-speed relation seems robust.

**Prediction** - For a perceptron problem with $n$ attributes (drawn from the same distributions) and a desired accuracy, the learning-speed relation tells how many examples are needed such there is a high probability the goal accuracy will be achieved.

**Measuring the Effect and Value of Knowledge** - Suppose background knowledge can be used to reduce the number of attributes by 30%. The learning relation predicts that this will reduce the number of examples needed by ~30%.

## 4.2 Backpropagation

The backpropagation experiment looked at problem properties $n_i$, the number of irrelevant attributes, and $n_r$, the number of relevant attributes. Examples were selected from $\{0, 1\}^{n_i + n_r}$ according to the uniform distribution. Potential targets were layered networks with $n_r$ input units, two internal units, and one output unit. The other $n_i$ inputs were ignored. The weights of the target networks were selected from the interval $[-1.0, 1.0]$ according to the uniform distribution. Each potential target was tested on randomly selected examples. Potential targets that classified fewer than 25% or more than 75% of the examples as positives were discarded.

Learning problems with $n_i = 0, 1, 2, 3, 5$ and 10 and $n_r = 1, 2, 3, 5$ and 10 were generated, 10 problems for each pair $< n_i, n_r >$. The learner was a backpropagation network with $n_i + n_r$ input units, 2 internal units, and 1 output units. It was given example training sets of size $m = 1, 5, 10, 25,$ and 50.. The learning network was allowed to run to convergence or for 500 epoches, whichever came first.

Multiple linear regression produced this learning-measure function:

$D(backpropagation_typical, n_i, n_r) = T(\bar{d}, s^2, df)$,

where $\bar{d}$, the mean, is $1.04 \times 10^{-2} n_i + 2.76 \times 10^{-2} n_r + -3.21 \times 10^{-2}$,

$s^2$, the variance, is $1.98 \times 10^{-2} \begin{bmatrix} 1 \\ n_i \\ n_r \end{bmatrix} \begin{bmatrix} 51.4 \times 10^{-4} & -4.27 \times 10^{-4} & -5.51 \times 10^{-4} \\ -4.27 \times 10^{-4} & 1.22 \times 10^{-4} & 0.0 \\ -5.51 \times 10^{-4} & 0.0 & 1.31 \times 10^{-4} \end{bmatrix} [1 \quad n_i \quad n_r]$

and $df$, the number of degrees of freedom, is 747.

$R^2$, the coefficient of determinism, is 0.31. The $F$ –test statistic (169.37) and the amount of data indicate that the significance is very high ($\alpha < 0.0001$). Linear regression on the average value of $d$ for each pair of $< n_i, n_r >$ resulted in a coefficient of correlation of 0.93 (figure 5b). Figure 6b shows that at point $< n_i = 4, n_r = 7 >$ the average of 10 learning curves is within the bounds predicted.

**Quantifying the Effect of Irrelevant Attributes** - If information about the number of irrelevant attributes is not available, this learning-speed relation will not be able to make useful predictions. This learning-speed relation can still, however, be used to better understand the backpropagation algorithm. Specifically, it quantifies backpropagation's sensitivity to irrelevant attributes. It shows that for this distribution of learning problems backpropagation is only 38% as sensitive to irrelevant attributes as it is to relevant attributes.

## 4.3 CSC

CSC is a decision-list learner. The CSC experiment measured $< n, l, e >$, the number of disjuncts per decision class, the length of the decision list, and the amount of noise (for details of the generation process see [Kadie, 1990]).

Multiple linear regression produced this learning-measure function:

$D(csc_typical, n, l, e) = T(\bar{d}, s^2, df)$

where $\bar{d}$, the mean, is $1.34 \times 10^{-1} n + 1.19 \times 10^{-1} l + 6.02 \times 10^{-1} e + -6.36 \times 10^{-1}$,

$s^2$, the variance, is

$$5.43 \times 10^{-1} \begin{bmatrix} 1 \\ n \\ l \\ e \end{bmatrix} \begin{bmatrix} 2.61 \times 10^{-3} & -3.57 \times 10^{-4} & -1.65 \times 10^{-4} & -1.71 \times 10^{-3} \\ -3.57 \times 10^{-4} & 1.19 \times 10^{-4} & 4.74 \times 10^{-32} & 3.73 \times 10^{-17} \\ -1.65 \times 10^{-4} & 4.74 \times 10^{-32} & 2.92 \times 10^{-5} & 1.28 \times 10^{-17} \\ -1.71 \times 10^{-3} & 3.73 \times 10^{-17} & 1.28 \times 10^{-17} & 1.00 \times 10^{-2} \end{bmatrix} [1 \quad n \quad l \quad e]$$

and $df$, the number of degrees of freedom, is 3146.

$R^2$, the coefficient of determinism, is 0.29. The $F$ –test statistic (421.66) and the amount of data indicate that the significance is very high ($\alpha < 0.0001$). Linear regression on the average value of $d$ for each triple of $< n, l, e >$ resulted in a coefficient of correlation of 0.86 (figure 5c). Figure 6c shows that at point $< n = 2, l = 5, e = 7\% >$ the average of 10 learning curves is within the bounds predicted.

**Measuring the Value of Noise Reduction** - With this model the value of noise reduction can be quantified. For example, reducing the noise level 50% would mean approximately $\frac{0.67 \times 10^{-1} n + 1.19 \times 10^{-1} l + 6.02 \times 10^{-1} e + -6.36 \times 10^{-1}}{1.34 \times 10^{-1} n + 1.19 \times 10^{-1} l + 6.02 \times 10^{-1} e + -6.36 \times 10^{-1}}$ % fewer examples would be needed to learn.

## 4.4 Learning with Constructive Induction

In the constructive induction experiments, examples were selected according to the uniform distribution. Targets were randomly-generated DNF expressions designed to classify about 50% of the examples as positives. The measured problem property was $l$, the number of literals in the DNF expression.

Learning problems with $l = 1, 4, 15$ and 44 were generated, 10 problems for each $l$. Two learning algorithms were used. The first was ID3 with no pruning. The second was the FRINGE, a feature-construction program. Both tried to learn with $m = 1, 5, 10, 25, 50$, and 200 examples.

Multiple linear regression produced this learning-measure functions:

$\mathrm{D}(dnf_typical, id3, l) = T(\bar{\mathrm{d}}_{id3,} \mathrm{s}^2_{id3,} df_{id3})$

where $\bar{\mathrm{d}}id3$, the mean, is $3.43 \times 10^{-2} l + -5.50 \times 10^{-2}$,

$\mathrm{s}^2_{id3}$, the variance, is $2.09 \times 10^{-2} \begin{bmatrix} 1 \\ l \end{bmatrix} \begin{bmatrix} 7.86 \times 10^{-3} & -2.31 \times 10^{-4} \\ -2.31 \times 10^{-4} & 1.44 \times 10^{-5} \end{bmatrix} [1 \quad l]$ ,

and $df_{id3}$, the number of degrees of freedom, is 178.

and

$\mathrm{D}(dnf_typical, fringe, l) = T(\bar{\mathrm{d}}_{fringe,} \mathrm{s}^2_{fringe,} df_{fringe})$

where $\bar{\mathrm{d}}fringe$, the mean, is $2.63 \times 10^{-2} l + -3.65 \times 10^{-2}$,

$\mathrm{s}^2_{fringe}$, the variance, is $2.09 \times 10^{-2} \begin{bmatrix} 1 \\ l \end{bmatrix} \begin{bmatrix} 7.86 \times 10^{-3} & -2.31 \times 10^{-4} \\ -2.31 \times 10^{-4} & 1.44 \times 10^{-5} \end{bmatrix} [1 \quad l]$ ,

and $df_{fringe}$, the number of degrees of freedom, is 178.

$R^2_{id3} = 0.33833$. $R^2_{fringe} = 0.36877$. $F - \mathrm{test}_{id3} = 91.017$.$F - \mathrm{test}_{fringe} = 103.99$. This indicates that the significance of both functions is very high ($\alpha < 0.0001$). Linear regression on the average value of $d_{id3}$ and $d_{fringe}$ for each $l$ resulted in a coefficients of correlations greater than .99 (figure 5d).

**Measuring the Value of Feature Construction (a Type of Constructive Induction**) - These models permit the value of FRINGE's feature construction to be quantified. Specifically, over this distribution of problems, the value of FRINGE's feature construction decreases as the complexity of these targets increases. For example, the learning-measure function reports that FRINGE needed 5.8% fewer examples when $l = 15$ and 0.9% fewer examples when $l = 44$.

# 5 Summary and Conclusion

This paper described the effective dimension, a new learning measure. It contributes a formal definition and a methodology for estimating the effective dimension of any learning problem. Five experiments illustrated the uses of effective-dimension analysis. These were:

- comparing empirical, average case to theoretical, worst case,
- prediction,
- measuring the effect and value of knowledge,
- quantifying the effect of irrelevant attributes,
- measuring the value of noise reduction, and
- measuring the value of feature construction (a type of constructive induction)

Future work should be done on finding additional applications; for example, using effective-dimension analysis to help set learning-program parameters. In addition, more learners (perhaps even humans) and learning properties should be analyzed. Also, the connection between learning measures, bias-strength measure, and complexity [Haussler, 1988] should be explored.

Two limitations of effective-dimension measure should be kept in mind. First, it is only as good as the effective-dimension assumption. The reasonableness of this assumption must always be evaluated. Second, the effective-dimension measure cannot escape context. If the appropriate learning-problem properties are not identified, learning measure will not be able to link the problem properties to the learning speed.

Despite these limitations, the effective-dimension measure offers important benefits. Like the VC dimension, the effective dimension relates directly to learning speed. In addition, because the effective dimension can be determined empirically, it can be used to measure average-case performance for a wide range of learning problems.